\pdfoutput=1
\documentclass[runningheads]{llncs}

\usepackage{eccv}

\usepackage{eccvabbrv}

\usepackage{graphicx}
\usepackage{booktabs}
\usepackage{amsmath}
\usepackage{tikz}
\usetikzlibrary{positioning, arrows.meta}
\usepackage{comment}
\usepackage{stfloats}

\definecolor{stageblue}{RGB}{70,130,180}
\definecolor{stageorange}{RGB}{210,140,60}
\definecolor{stagegreen}{RGB}{80,140,80}
\definecolor{stagecyan}{RGB}{70,160,180}
\definecolor{stagelightblue}{RGB}{130,170,210}

\usepackage[accsupp]{axessibility}  %

\usepackage{hyperref}

\usepackage{orcidlink}

\DeclareRobustCommand{\OURS}{WorldMesh}
\newcommand{\CR}[1]{#1}

\begin{document}

\title{\OURS: Generating Navigable Multi-Room 3D Scenes via Mesh-Conditioned Image Diffusion}

\titlerunning{\OURS}

\author{Manuel-Andreas Schneider\,\orcidlink{0009-0009-6778-1191} \and Angela Dai\,\orcidlink{0000-0002-6241-8782}}

\authorrunning{M.-A.~Schneider, A.~Dai}

\institute{Technical University of Munich, Germany\\
\CR{\email{\{manuel.schneider, angela.dai\}@tum.de}}\\
\CR{Code \& data:}~\url{https://mschneider456.github.io/world-mesh/}}

\maketitle

\begin{figure}
\begin{center}
\includegraphics[width=0.99\textwidth]{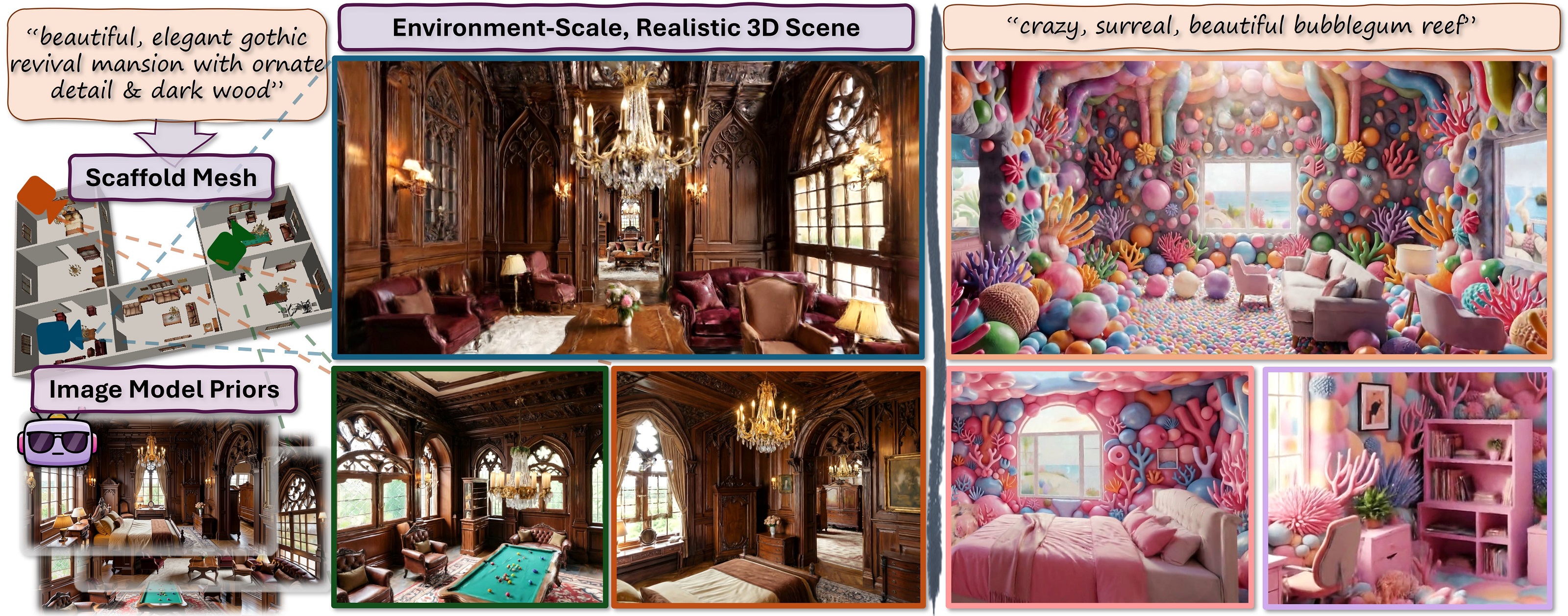}
\end{center}
\vspace{-0.5cm}
\caption{\OURS{} tackles environment-scale 3D scene synthesis by decoupling this complex problem into structure and appearance. From  an input text prompt, we first construct a mesh scaffold that establishes the output scene's layout and geometric structure. This mesh is then used as scaffold for conditioned image synthesis to encourage multi-view consistency across both local (e.g., around objects) and global (e.g., across rooms) scales. Synthesized views are optimized into 3D gaussian splats representing a navigable 3D world through view synthesis.}
\label{fig:teaser}
\end{figure}

\begin{abstract}
Recent progress in image and video synthesis has inspired their use in advancing 3D scene generation. However, we observe that text-to-image and -video approaches struggle to maintain scene- and object-level consistency beyond a limited environment scale \CR{without a persistent, explicit geometric representation}.
We thus present a geometry-first approach that decouples this complex problem of large-scale 3D scene synthesis into its structural composition, represented as a mesh scaffold, and realistic appearance synthesis, which leverages powerful image synthesis models conditioned on the mesh scaffold.
From an input text description, we first construct a mesh capturing the environment's geometry (walls, floors, etc.), and then use image synthesis, segmentation and object reconstruction to populate the mesh structure with objects in realistic layouts. 
This mesh scaffold is then rendered to condition image synthesis, providing a structural backbone for consistent appearance generation. 
This enables scalable, arbitrarily-sized 3D scenes of high object richness and diversity, combining robust 3D consistency with photorealistic detail. We believe this marks a significant step toward generating truly environment-scale, immersive 3D worlds.
\end{abstract}

\section{Introduction}
\label{sec:intro}

Photorealistic 3D environments are central to computer vision and graphics, with applications spanning virtual reality, video games,
architecture visualization, and interior design. Creating such environments
manually requires significant artistic effort: modeling, texturing, and
lighting a single room can take days to weeks for a skilled 3D artist, and scaling to
multi-room apartments or larger buildings multiplies this effort dramatically.

Recent advances in generative modeling, in particular in text-to-image and -video generation models \cite{rombach2022highresolution,lipman2023flow,esser2024scaling,blattmann2023stablevideo}, have inspired new approaches to 3D scene synthesis.
Several recent approaches have leveraged the flexible, photorealistic synthesis capabilities of these image and video generative models to outpaint or synthesize various images of a scene, which can then be used to reconstruct an output 3D scene representation \cite{hoellein2023text2room,schult2024controlroom3d,schneider2025worldexplorer}.
While these methods can produce impressive single-room synthesis, the very lack of explicit 3D structure in these 2D-based synthesis models, which enables their flexible, powerful synthesis, fundamentally hinders their applicability to world-level, environment-scale scenes.
In particular, the absence of 3D structure makes it difficult for such methods to maintain  coherent appearance across diverse viewpoints around objects, especially close-up or tightly rotating views.
Additionally, in large, multi-room environments these limitations often manifest as inconsistencies across views and rooms, making it largely intractable to scale such methods to arbitrarily large, complex environments.

We thus propose \OURS{} \CR{(\cref{fig:teaser})}, a geometry-first approach that decouples the complex task of large-scale 3D scene synthesis into its fundamental challenges:
\emph{global spatial consistency} and \emph{local photorealistic appearance}.
From an input text prompt, we construct our mesh scaffold representing the output environment's layout. The mesh scaffold is then populated with objects based on image synthesis to suggest object placement and appearance. 
This enables our scaffolding of image diffusion onto this explicit 3D mesh that maintains inherent spatial coherence  across arbitrary camera poses (e.g., close up views, views transitioning across rooms), while
preserving the image diffusion model's generative flexibility for local appearance. 
The generated images can be reconstructed into a navigable 3D scene using 3DGS~\cite{kerbl2023gaussian} optimization anchored to the mesh scaffold geometry, maintaining geometric fidelity, multi-room connectivity, and visually diverse, realistic content.
This principled separation enables multi-room generation at a scale that remains
intractable for purely pixel-space models.

We make the following contributions:
\begin{itemize}
  \item We present the first approach for scalable, multi-room 3D synthesis from text prompts, supporting diverse visual themes, by constructing a mesh scaffold to represent scene geometry and using it to anchor mesh-conditioned appearance generation.
  Our approach scales linearly with the number of rooms.
  \item We introduce mesh-guided appearance synthesis, where the rendered scaffold provides a structural backbone signaling depth, object geometry, and textures, to be used as condition for consistent image synthesis. 
  We reconstruct objects as part of the mesh scaffold based on synthesized images, in order to preserve realistic layouts and high object richness, demonstrating that mesh-conditioning can scale photorealistic 3D scene generation to arbitrarily large, multi-room environments with robust 3D consistency.
\end{itemize}

\section{Related Work}
\label{sec:related}

\paragraph{Text-to-Image/Video through Score-based Modeling.}
Score-based generative models have become the dominant paradigm for high-fidelity image and video synthesis. 
Early diffusion models~\cite{ho2020denoising,song2021scorebased} generate images by iteratively denoising a stochastic process in raw image pixel space, while latent diffusion models (LDMs)~\cite{rombach2022highresolution} perform this process in a compressed latent space, enabling high-quality generation at practical inference-time computational costs.
More recently, transformer-based approaches such as  Diffusion Transformers
(DiT)~\cite{peebles2023scalable}, along with flow matching~\cite{lipman2023flow} training, have improved  quality and sampling speed, as adopted by the Flux model
family~\cite{flux2024}. Additionally, controllable generation has been advanced through ControlNet~\cite{zhang2023controlnet}, which injects spatial conditions (e.g., depth maps) into frozen diffusion models, and IP-Adapter~\cite{ye2023ipadapter}, which enables image-prompt conditioning.
In addition to images, video generation has also seen remarkable progress through score-based generative modeling \cite{blattmann2023stablevideo,brooks2024video} as well as camera-conditioned control \cite{bahmani2024vd3d,bahmani2024ac3d,he2024cameractrl2}, though they typically require substantially more compute than image models.
Our approach leverages the generative power and relative efficiency of modern image synthesis models, but rather than relying on unconstrained image generation, we introduce explicit geometric conditioning with a mesh scaffold, enabling image synthesis to produce photorealistic appearance that remains 3D consistent across large-scale 3D scenes.

\paragraph{3D Scene Generation with Image and Video Generative Priors.}
Inspired by the success of image and video generative models, a growing body of work has explored leveraging these models for text-driven 3D generation.
DreamFusion~\cite{poole2023dreamfusion} pioneered a text-to-3D approach from 2D image generative models via score distillation into a NeRF~\cite{mildenhall2020nerf}, showing powerful object-based synthesis but struggling with the complexity of room-scale scenes. 
Subsequent methods have also proposed to generate many images of a scene in order to reconstruct the 3D scene geometry. 
Inpainting-based methods like Text2Room~\cite{hoellein2023text2room}, WonderJourney~\cite{yu2023wonderjourney}, and WonderWorld~\cite{yu2024wonderworld}  iteratively synthesize and fuse images to expand a camera trajectory. 
This can achieve impressive single-room results, but tends to accumulate geometric drift over long trajectories. 
Multi-view generative approaches, such as MVDiffusion~\cite{tang2024mvdiffusion} and CAT3D~\cite{gao2024cat3d}, improve view consistency through joint multi-view reasoning, and DreamScene~\cite{li2024dreamscene} uses panoramic diffusion with 3DGS, but these methods target single objects or bounded scenes. 

Video-based approaches including
ReconX~\cite{liu2024reconx}, DimensionX~\cite{sun2024dimensionx}, and
FlexWorld~\cite{chen2024flexworld} leverage video diffusion models as strong 3D priors for scene generation. While these methods can improve coherence, they require substantial GPU resources and remain difficult to scale beyond single-room environments.
\CR{While recent video models increasingly incorporate explicit geometric priors such as camera poses or point-cloud renderings, maintaining consistency over very long, multi-room trajectories remains challenging and computationally demanding.}
WorldExplorer~\cite{schneider2025worldexplorer}  relies on a camera-conditioned  generative model prior~\cite{zhou2024stablevirtualcamera} to synthesize new views in a scene, thereby constructing navigable 3D scenes.
However, because these approaches operate largely in pixel space without an explicit geometric scaffold, maintaining spatial consistency across diverse viewpoints remains challenging. In particular, they often struggle with viewpoints that move close to objects or traverse larger spatial extents (e.g., across multiple rooms), where the limited context of the generative model and the absence of explicit 3D structure can lead to inconsistent geometry and appearance.

Recently, several works have introduced layout guidance to help control the generative process. Methods such as ControlRoom3D~\cite{schult2024controlroom3d}, SceneCraft~\cite{hu2024scenecraft}, and SpatialGen~\cite{SpatialGen} rely on user-provided coarse 3D layouts (e.g., object bounding boxes) to help improve consistency. 
Our approach also observes that structural priors can help ground consistency; however, we generate the layout from the text prompt and instantiate it as an explicit mesh scaffold that enables more precise geometric anchoring for appearance synthesis.

\paragraph{3D Object Generative Models}
3D object generation has also seen remarkable advances, in particular through score-based generative modeling paradigms. 
Early diffusion-based methods explored diffusion modeling across various 3D latent shape representations \cite{vahdat2022lion,erkocc2023hyperdiffusion,zhang2025dnf,zhang20233dshape2vecset}.
Recently, TRELLIS~\cite{xiang2025structured} introduced a structured 3D latent representation for objects, which coupled with flow matching training, produces extremely compelling, high-fidelity 3D shapes from text or image inputs. 
SAM-3D-Objects~\cite{meta2025sam3dobjects} employs a similar 3D shape representation to reconstruct objects along with their poses to match an input image.
Such methods have shown powerful generative capabilities, but target single objects, which avoids challenges with arbitrary sizes, lack of canonicalization, and large resolutions required for scene-level modeling.
We also employ the generative capacity of such object generative models to help instantiate estimated object geometry into our constructed mesh scaffold, which is then used as a geometric anchor for photorealistic appearance synthesis.

\section{Method}
\label{sec:method}

\subsection{Overview}
\label{sec:method-overview}

\begin{figure}[t]
\begin{center}
\includegraphics[width=0.99\textwidth]{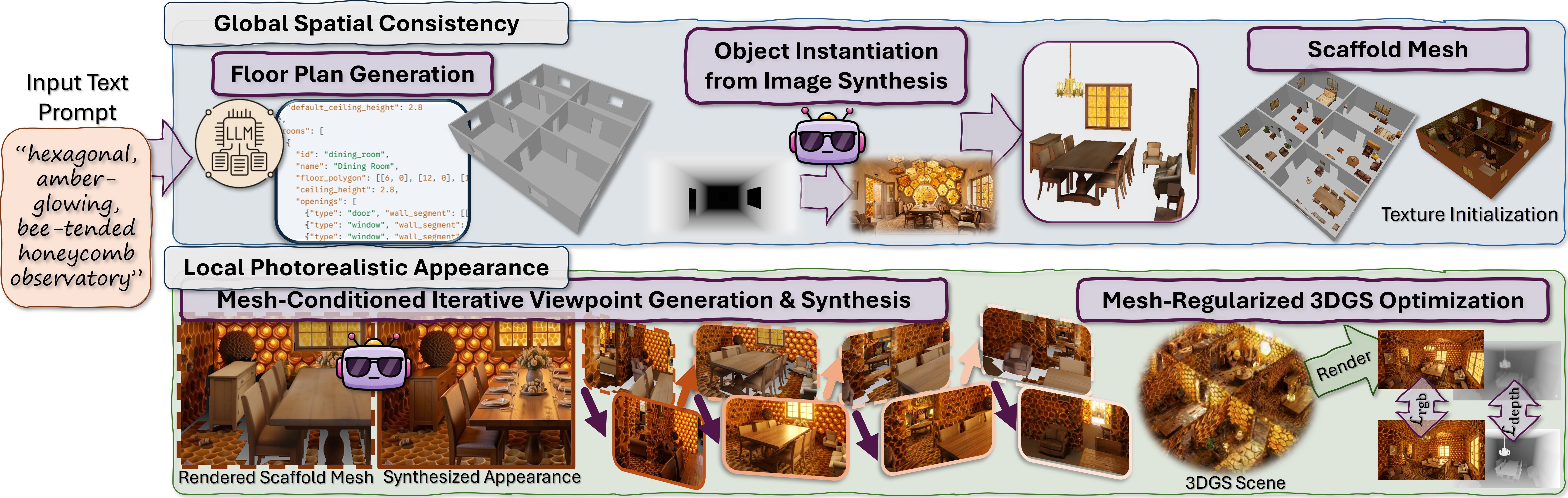}
\end{center}
\vspace{-0.5cm}
\caption{ Overview of \OURS. To generate a complex, multi-room 3D scene from a text prompt, we decompose this problem into first constructing the global scene structure as a mesh scaffold (top), and then using the scaffold mesh as anchor for realistic local appearance (bottom). 
The text prompt is used to generate a text-based floor plan, which we construct in 3D to use as depth conditioning for an image synthesis model $\Phi$, in order to reconstruct estimated 3D objects in each room.
The structural elements and 3D objects constitute the scaffold mesh $\mathcal{M}$\CR{; its wall textures are not pre-generated but accumulated on-the-fly, by projecting each synthesized view back onto $\mathcal{M}$ during the synthesis loop below}.
$\mathcal{M}$ then serves as a geometric anchor for iterative image synthesis using $\Phi$ to generate images $\{I_i\}$. Finally, the output scene $\mathcal{S}$ is optimized with geometry-regularized 3DGS, against both images $\{I_i\}$ and rendered depth from $\mathcal{M}$.
}
\label{fig:method}
\end{figure}

Our aim is to generate large, environment-scale, multi-room 3D scenes from a natural language text prompt input $\mathcal{T}$ (e.g., ``a cozy Scandinavian apartment''), with an output scene $\mathcal{S}$ represented as 3D gaussian splats \cite{kerbl2023gaussian}.
Large-scale multi-room scene generation is challenging: purely 2D or video-based diffusion produces photorealistic local detail but struggles to maintain long-range multi-view and multi-room consistency, particularly around objects and in complex layouts. To address this, we decouple the problem into a geometry-first approach: (i), first, we construct a mesh scaffold $\mathcal{M}$ that encodes the 3D geometric structure of the scene, and (ii) then, we generate photorealistic appearance anchored to this scaffold, enabling both local and global coherence \CR{(\cref{fig:method})}.

\subsection{Mesh Scaffold Construction}
\label{sec:mesh-generation}

To build the mesh scaffold, we first generate from the input text prompt $\mathcal{T}$ a floor plan layout $L$, represented in JSON text format, to characterize the output 3D scene.
The layout $L$ is generated with Claude Opus 4.6 \cite{anthropic2026opus46} and specifies structural scene metadata such as wall thickness, ceiling height, and room definitions, including floor polygons and openings (doors or passageways). Candidate layouts are sampled and validated against a set of spatial validity rules (e.g., no windows between rooms), and sampling continues until a layout is found that satisfies these coarse spatial validity constraints.
This forms the blueprint for the structural mesh scaffold $\mathcal{M}$.

Based on $L$, we first instantiate the 3D structure of the scene as $\mathcal{M}_\text{struct}$.
Each room is defined by a 2D floor polygon, openings, as well as ceiling height.
To construct $\mathcal{M}_\text{struct}$, each floor polygon is first inset by the wall thickness to define wall cross-sections, which are then extruded vertically to the ceiling height, with shared walls between adjacent rooms represented at half thickness. Matching edges between rooms are detected within a 0.01\,m tolerance to ensure continuous walls, and door and window openings are carved using boolean subtraction, with automatic propagation to adjacent rooms sharing the wall. Finally, thin planar extrusions are added to form floors and ceilings, completing the structural mesh $\mathcal{M}_\text{struct}$. $\mathcal{M}_\text{struct}$ then provides a geometrically consistent layout for the scene and can be exported as a GLB file.

\paragraph{Camera Viewpoint Generation.} 
A set of cameras $\{c_i^r\}$ is generated for each room $r$ and reused across all method stages (i.e., object instantiation and final appearance synthesis).
Two \emph{bootstrap cameras}, $c_0^r$ and $c_1^r$, are placed at eye-level at opposite wall midpoints, on the shorter walls of the rectangular rooms, together spanning all four walls.
Approximately 16 \emph{perimeter cameras} at eye-level are uniformly distributed along the room walls with a small wall offset \CR{(spaced by arc length, with any perimeter camera within $10^\circ$ of a bootstrap view pruned to avoid redundancy)}, each looking toward the room center.
Eight \emph{overhead cameras} follow the same perimeter layout but are positioned higher up and look downward for wider spatial coverage.
All cameras are nudged away from placed objects to avoid collisions.
During initial object placement (\cref{sec:object-extraction}), only the first coverage camera $c_0^r$  is rendered. During final image generation for 3DGS optimization the full camera set is used.

\subsubsection{Object Instantiation}
\label{sec:object-extraction}

While LLM-generated layouts provide reliable structural organization, their textual nature can lead to  stale or underspecified object arrangements, lacking the richness and diversity characteristic of real indoor environments. 
In contrast, modern image synthesis models naturally produce complex and visually coherent object configurations, capturing implicit priors about furniture placement, clutter, and stylistic composition. We therefore leverage image synthesis to populate each room with objects, using the previously constructed floor plan scaffold $\mathcal{M}_\text{struct}$ as geometric guidance.

For each room, we render $\mathcal{M}_\text{struct}$ from an initial coverage camera pose $c_0$ located at the wall midpoint of the smallest side of the rectangular room for maximal coverage. This produces a depth map $D_0$ which is used to condition a photorealistic image generation model (Flux2-Klein~\cite{flux2klein2025}). 
More concretely, generation is guided jointly by $D_0$, a text prompt combining the room type, global text theme $\mathcal{T}$, and visible architectural context information from $L$ (e.g., door or window location). The architectural context is used to inform the model about the nature of depth maps since it might otherwise interpret the provided openings visible through the depth map as something different (e.g., a TV).
The resulting image $I_o^\text{obj}$ then establishes the visual style and an initial object layout consistent with the scene scaffold.

We then extract individual 3D object instances from the generated images using SAM3~\cite{meta2025sam3}. \CR{In our pipeline these masks are obtained interactively, via a few point prompts per object in SAM3; the segmentation could alternatively be run automatically from a list of text prompts (see appendix).}
Each segmented object is reconstructed into a 3D representation using SAM-3D-Objects~\cite{meta2025sam3dobjects}, which generates both object meshes in a canonical coordinate system along with their object transforms to the camera coordinate system of the image. 
As SAM-3D-Objects makes independent predictions per object, we encourage physical plausibility of the outputs by aligning their oriented bounding boxes with the gravity direction of the scene and projecting objects onto the closest support surface below them (or above them in the case of lamps, etc.).
The output mesh with composed objects instantiated from the image model prior for all rooms is denoted  $\mathcal{M}_\text{geo}$, which contains object textures and geometry from SAM-3D-Objects as well as structural floor plan geometry from the layout $L$.

\subsubsection{Mesh Scaffold Texturing}
\label{sec:conditioning}

\CR{We also texture the structural elements of the scaffold (e.g., walls). This wall texturing is not a separate stage that pre-textures the mesh before image generation; it is performed jointly with appearance synthesis, projecting each generated view back onto $\mathcal{M}_\text{geo}$ as it is produced.} As multi-view consistency of wall appearance between different camera views can be a strong challenge, instead of relying on a 2D image generative model to implicitly preserve color consistency, we explicitly propagate appearance information across views by initializing structural textures through projective texture accumulation. \CR{Because the two bootstrap views of each room observe opposite walls and together span all four walls, these first two generations already texture a large portion of the structural surfaces, while the remaining views progressively fill in the rest.}

Once an image is generated from a given camera pose $c_i$, its pixel colors are projected back onto visible wall surfaces of $\mathcal{M}_\text{geo}$ using projective texturing~\cite{heckbert1986survey}. 
Given the known camera intrinsics $(f_x, f_y, c_x, c_y)$ and camera-to-world transform $c_i$, each mesh vertex $\mathbf{v}_\text{world}$ is projected to its pixel coordinates via the
world-to-camera transform $c_i^{-1}$. The UV coordinates for texture mapping are then $(u = \frac{p_x}{W}, \quad v = 1 - \frac{p_y}{H})$, where $W$ and $H$ are image dimensions and the $v$-flip follows the glTF convention.

Note that this \CR{accumulated} mesh texturing \CR{provides} a strong conditioning signal, but does remain incomplete in various regions due to many object occlusions. 
The accumulated projections can also introduce seams, and object textures inherited from SAM-3D-Objects are often relatively coarse.
Our final appearance synthesis addresses this by enriching the full scene with fine-scale photorealistic detail.

\subsection{Mesh Anchored Photorealistic Appearance Synthesis}

The constructed mesh scaffold $\mathcal{M}$ now represents the underlying geometric structure of the scene (walls, floors, ceilings, openings, objects).
This then serves as a guide for synthesizing photorealistic appearance anchored to $\mathcal{M}$ to enable coherent appearance across views and rooms.
Specifically, we employ an image synthesis model $\Phi$ conditioned on depth and color renderings  of $\mathcal{M}$, using the scaffold to enforce structural fidelity while allowing the model to generate realistic materials, lighting, and fine-grained visual detail. In this way, appearance is not generated freely in 2D, but is instead anchored to the underlying 3D geometry.

\paragraph{Viewpoint Generation for Appearance Synthesis.}

To synthesize the appearance, we use the set of generated camera viewpoints $\{c_i\}$ to create an output set of multi-view images from a 2D image synthesis model  $\Phi$.  
Note that generating all views independently, even when conditioned on the mesh  $\mathcal{M}$, would introduce inconsistencies in color, lighting, and materials due to the 2D-only nature of $\Phi$. 
We thus design a viewpoint generation strategy that expands coverage progressively while propagating style through the output scene.

For each room $r$, we begin with the two previously defined bootstrap cameras, $c_0^r$ and $c_1^r$, to establish the room's global appearance. As $c_0^r$ and $c_1^r$ have been placed at eye-level at opposite wall midpoints, they cover all four walls. These two views enable broad coverage over the room $r$, which stabilizes  appearance conditioning  for later view synthesis, reducing the risk of stylistic drift.

After bootstrapping with $c_0^r$ and $c_1^r$, the remaining camera poses $\{c_i^r\}$ are sampled iteratively.
The next camera is selected by greedy nearest-neighbor rotational similarity from the current camera set.
At each step, the candidate with the highest quaternion similarity to any
previously generated camera is selected:
\begin{equation}
  \text{sim}(q_1, q_2) = |\hat{q}_1 \cdot \hat{q}_2|,
  \label{eq:quat-similarity}
\end{equation}
where $\hat{q}$ denotes a unit quaternion. This creates a gradual outward spiral where each camera has a maximally similar style reference to encourage consistent appearance synthesis. Following generation, the image is projected back to the texture map of $\mathcal{M}$, to obtain the updated $\mathcal{M}_i$.
This enables the conditioning signal for the future camera viewpoints to reflect the current appearance information. 

\paragraph{Mesh-Conditioned Image Synthesis.}

To guide image synthesis, we employ a conditioning signal that explicitly separates structural geometry from object appearance while incorporating the estimated wall textures. Note that in the following, we discard the room index $r$ for simplicity.

To construct this, for each camera pose $c_i$ in the generated viewpoints, an image $X_i$ is rendered where  objects are rendered with their texture, structural elements are rendered with the alpha-blended wall textures where available, and all other pixels are assigned to their grayscale depth. 
For the first image conditions $X_0$ and $X_1$, corresponding to the bootstrap cameras $c_0$ and $c_1$ in the first room synthesized (no previously generated conditions), the object and wall textures originate from SAM-3D-Objects~\cite{meta2025sam3dobjects} and our wall texturing, respectively. Each following view (including for additional rooms) then uses the updated texture information from previously synthesized views.

For each camera $c_i$, we select as style reference the previously generated image $I_{k(i)}$ whose camera pose is rotationally closest, $k(i) = \arg\max_{j < i}\, \mathrm{sim}(q_j, q_i)$ (\cref{eq:quat-similarity}), providing appearance guidance to the synthesis model.
Each output image is then synthesized as $I_i = \Phi(X_i,\, I_{k(i)},\, \mathcal{T})$.

\paragraph{Image Verification}

Even with strong geometric conditioning, image synthesis models may occasionally violate structural constraints. 
To ensure that a generated image $I_i$ from $\Phi$ maintains consistency with the geometry of the mesh scaffold $\mathcal{M}$, we validate $I_i$ by comparing its implied geometry to that of $\mathcal{M}$.

As $I_i$ has been created by an image synthesis model, its underlying geometry is unknown. 
We thus instead estimate a depth map $D_i'$ with a state-of-the-art depth estimation model \cite{bochkovskii2024depthpro}.
Rather than relying on raw depth estimates, which may contain noise and some depth-scale ambiguity errors, we extract depth edges $E'_i$ through Canny edge detection, which captures the major geometric structures in the scene. 
These edges are then compared against edges $E^\mathcal{M}_i$  extracted from the depth maps rendered from $\mathcal{M}$ to evaluate the structural fidelity of $I_i$.
Specifically, the validation is measured as depth edge recall:
\begin{equation}
  \text{Recall} =
    \frac{|E^\mathcal{M}_i \cap
      \text{dilate}(E'_i, \delta)|}
         {|E^\mathcal{M}_i|},
  \label{eq:edge-recall}
\end{equation}
where $E^\mathcal{M}_i$ and $E'_i$ are the Canny-detected edge pixel sets and $\delta = 10$\,px is the dilation tolerance. \CR{Canny edges are extracted with hysteresis thresholds $(0.1, 0.3)$ of the normalized depth range and a $3{\times}3$ Sobel aperture.} Edge recall measures only \emph{missing} structural edges (false negatives), not extra edges from furniture (false positives), as we allow the image model $\Phi$ to imagine  plausible objects not present in $\mathcal{M}$. A generated image passes if its recall exceeds the threshold, and is otherwise regenerated with a new seed.

\paragraph{Geometry-Regularized 3DGS Scene Reconstruction.}

For all successfully validated views $\{I_i\}$, we optimize a 3DGS representation of the output scene $\mathcal{S}$ using the synthesized images $I_i$, their camera poses $c_i$, and depth maps $D_i$ rendered from the mesh scaffold $\mathcal{M}$.
The Gaussian splats are initialized from a point cloud obtained by back-projecting the mesh depth maps $D_i$, providing a strong geometric prior for optimization.
The training loss combines photometric reconstruction with depth regularization:
\begin{equation}
  \mathcal{L} = (1 - \lambda_\text{s}) \, \mathcal{L}_1(I_i, \hat{I}_i) \;+\; \lambda_\text{s} \, \mathcal{L}_\text{DSSIM}(I_i, \hat{I}_i) \;+\; \lambda_\text{d} \, \|D_i - D^\mathcal{S}_i\|_1,
  \label{eq:3dgs-loss}
\end{equation}
where $\hat{I}_i$ and $D^\mathcal{S}_i$ are the image and depth rendered from $\mathcal{S}$ at camera $c_i$, $\mathcal{L}_\text{DSSIM} = 1 - \text{SSIM}$, and we set $\lambda_\text{s} = 0.2$, $\lambda_\text{d} = 0.7$.
The depth term prevents geometric drift, a common failure mode in 3DGS under sparse views, and ensures that $\mathcal{S}$ preserves the architectural structure of~$\mathcal{M}$.

\section{Results}
\label{sec:experiments}

\begin{figure}[t]
\begin{center}
\includegraphics[width=0.83\textwidth]{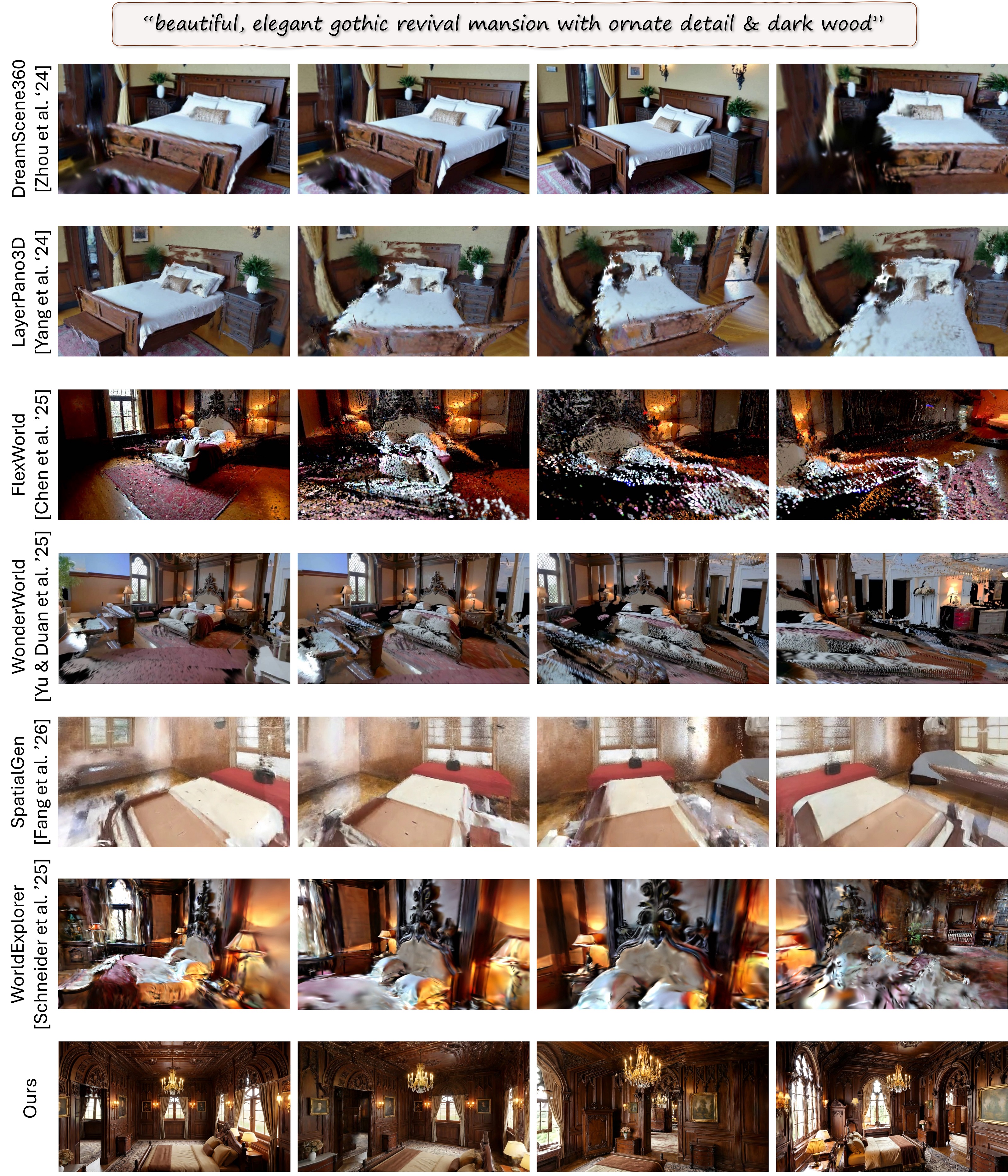}
\end{center}
\vspace{-0.4cm}
\caption{Qualitative comparison with baselines. We compare with state-of-the-art methods for 3D scene generation leveraging image, panorama, and video generative model priors \cite{zhou2024dreamscene360,yang2025layerpano3d,chen2024flexworld,yu2024wonderworld,SpatialGen,schneider2025worldexplorer}. Since these methods focus on a single-room setting, we show our results focusing on one of our generated rooms within our multi-room generation. Most baselines lack explicit 3D structure, leading to view inconsistencies, particularly for challenging viewpoints close to objects (as views are typically synthesized with wider range for easier scene coverage and consistency). SpatialGen~\cite{SpatialGen} mitigates this with 3D box-based layout conditioning, but struggles to maintain realistic local detail. In contrast, our mesh scaffold anchors synthesis geometrically, enabling much stronger multi-view consistency and more realistic appearance across diverse views.
}
\label{fig:results_comparison}
\end{figure}

\paragraph{Implementation Details}

We generate various large-scale scenes with different number of rooms, room dimensions and structural layout, each with diverse visual themes. 
All of our outputs were generated on a single NVIDIA RTX A5000 GPU (24\,GB VRAM).

For the images synthesized for object placement we use the Flux2-Klein model locally as the image synthesis model $\Phi$, as we found it the most capable of producing diverse, complex visual results. The images synthesized for object placement use a field of view of $90^\circ$ for wider coverage; all other images are synthesized with $60^\circ$ field of view. %

Once objects have been placed in the mesh, we generate \CR{$\approx$}24 images per room, each of resolution $1376 \times 768$. For the first image conditioned on the untextured mesh with textured objects, we use Flux2-Klein locally. For subsequent image generations we use the Nano \CR{Banana} Pro \CR{(NB~Pro)} \cite{google2025nanobanana} model through their API, finding that it is most faithful to the provided mesh priors. \CR{Our pipeline is nonetheless backbone-agnostic: it also runs fully locally and open-source with Flux2-klein (9B) on a single RTX~A5000, and the layout LLM can likewise be swapped (see the supplementary material).}

\paragraph{Baselines.}
We compare against multiple types of text-to-3D scene, as well as layout guided 3D scene generation methods.

\begin{itemize}
  \item \emph{Layout-guided 3D Indoor Scene Generation}: most related to our approach, we compare against SpatialGen \cite{SpatialGen}, which takes as input a 3D layout and a reference image to synthesize appearance and geometry from arbitrary viewpoints. %
  Since our method takes only a text prompt as input, we provide SpatialGen our generated room layout and object locations as layout input, along with our initial object placement image as a reference image.
  \item \emph{Autoregressive Video Diffusion for Navigable 3D Scenes}: we compare against WorldExplorer~\cite{schneider2025worldexplorer} as the current open-source state-of-the-art for text-driven world generation along pre-defined camera trajectories. We also compare with FlexWorld~\cite{chen2024flexworld}, which generates $360^\circ$ scenes with video diffusion. Both are evaluated using the same text prompts as our method.
  \item \emph{Iterative T2I Lifting}: we compare with WonderWorld~\cite{yu2024wonderworld}, which constructs 3D scenes via render-refine-repeat using T2I models and monocular depth estimation. We provide one image generated with our text prompt as input to the model for comparison.
  \item \emph{Panorama-To-3D}: we compare with DreamScene360~\cite{zhou2024dreamscene360} and LayerPano3D~\cite{yang2025layerpano3d}, both of which first generate a $360^\circ$ panorama from text, which is used to reconstruct a 3DGS \cite{kerbl2023gaussian} scene.
\end{itemize}

\paragraph{Metrics.} For subjective scene quality, we compute CLIP-IQA+~\cite{wang2022exploring} and CLIP Aesthetic~\cite{schuhmann2022clip} scores on frames rendered along trajectories of the final 3DGS scenes. To assess 3D consistency at both environment and object levels, we conduct a perceptual study with 31 participants, who rate object consistency, scene-structure consistency, and overall quality on a 1--5 scale, and give binary preferences between each baseline and our method in randomized order. We evaluate challenging trajectories containing close-up views and rotations across objects.

\subsection{Qualitative Results}
\label{sec:qualitative}

\begin{figure}[t]
\begin{center}
\includegraphics[width=0.78\textwidth]{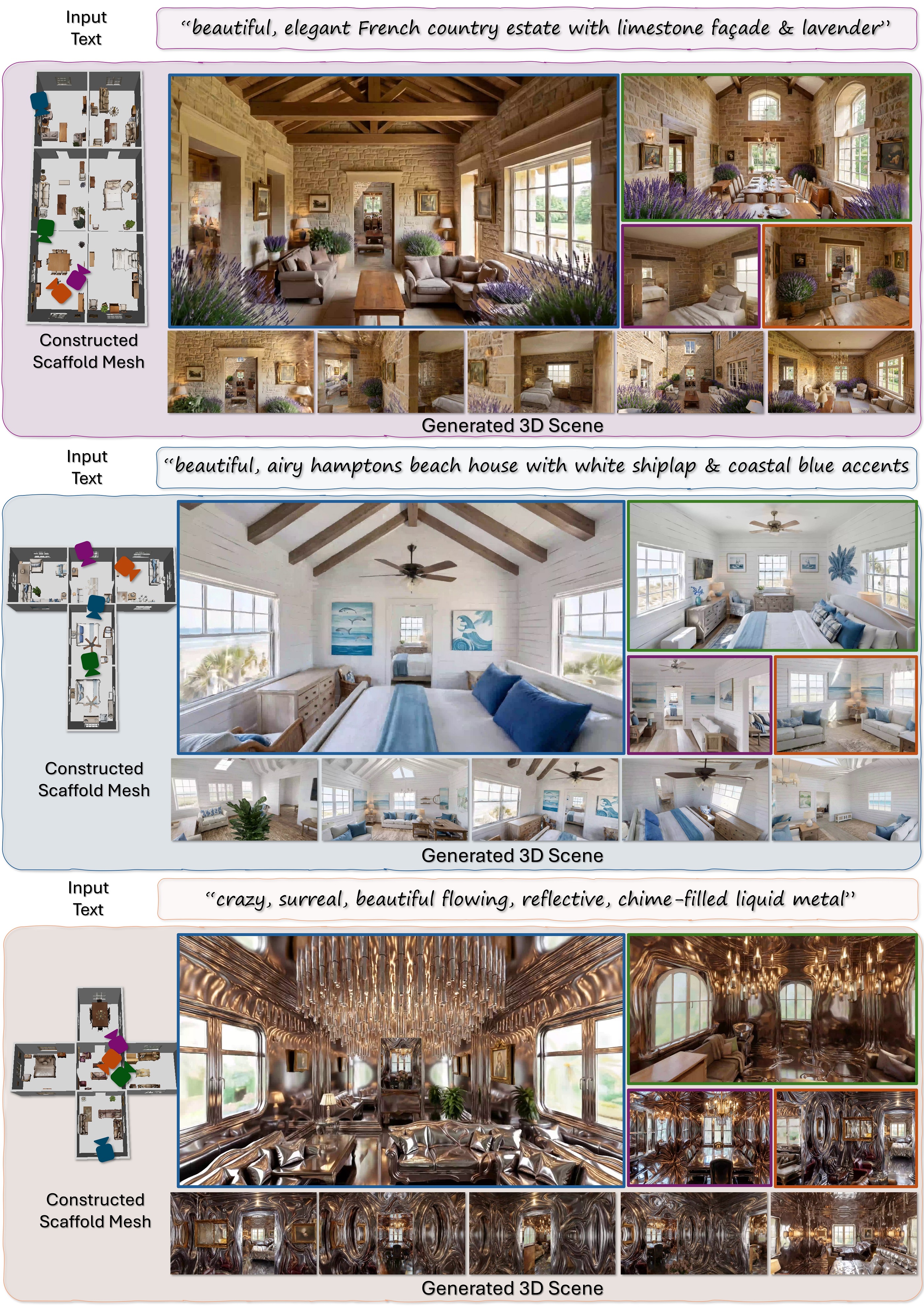}
\end{center}
\vspace{-0.5cm}
\caption{\OURS{} generates a diverse range of complex 3D scenes, varying both in spatial layouts and visual themes.
We show rendered views from different rooms within each multi-room generated environment, including transitional viewpoints between rooms, demonstrating that stylistic coherence and 3D consistency are maintained throughout.
}
\label{fig:results_ours}
\end{figure}

\cref{fig:results_comparison} shows a qualitative comparison for views rotating around a bed, highlighting the advantages of our mesh-anchored approach. 
Both WorldExplorer~\cite{schneider2025worldexplorer} and FlexWorld~\cite{chen2024flexworld} produce good results along predefined camera trajectories, but struggle with object consistency in close up views. 
Panorama-based methods like DreamScene360~\cite{zhou2024dreamscene360} and LayerPano3D~\cite{yang2025layerpano3d} generate high-fidelity $360^\circ$ views, but struggle to generate 3D-consistent output beyond their central perspective, causing occlusions and distortions. 
The iterative WonderWorld~\cite{yu2024wonderworld} builds up accumulated error from inaccurate depth estimates, while SpatialGen~\cite{SpatialGen} partially mitigates inconsistencies with box-based layout conditioning, but produces less realistic local detail.
In contrast, our mesh scaffold provides global spatial consistency, enabling reliable 3D-consistent synthesis even in close-up object views.

We  show additional results of our method across a range of realistic and surreal scene types and environments in \cref{fig:results_ours}, as well as in the appendix.

\begin{figure}[t]
\begin{center}
\includegraphics[width=0.8\textwidth]{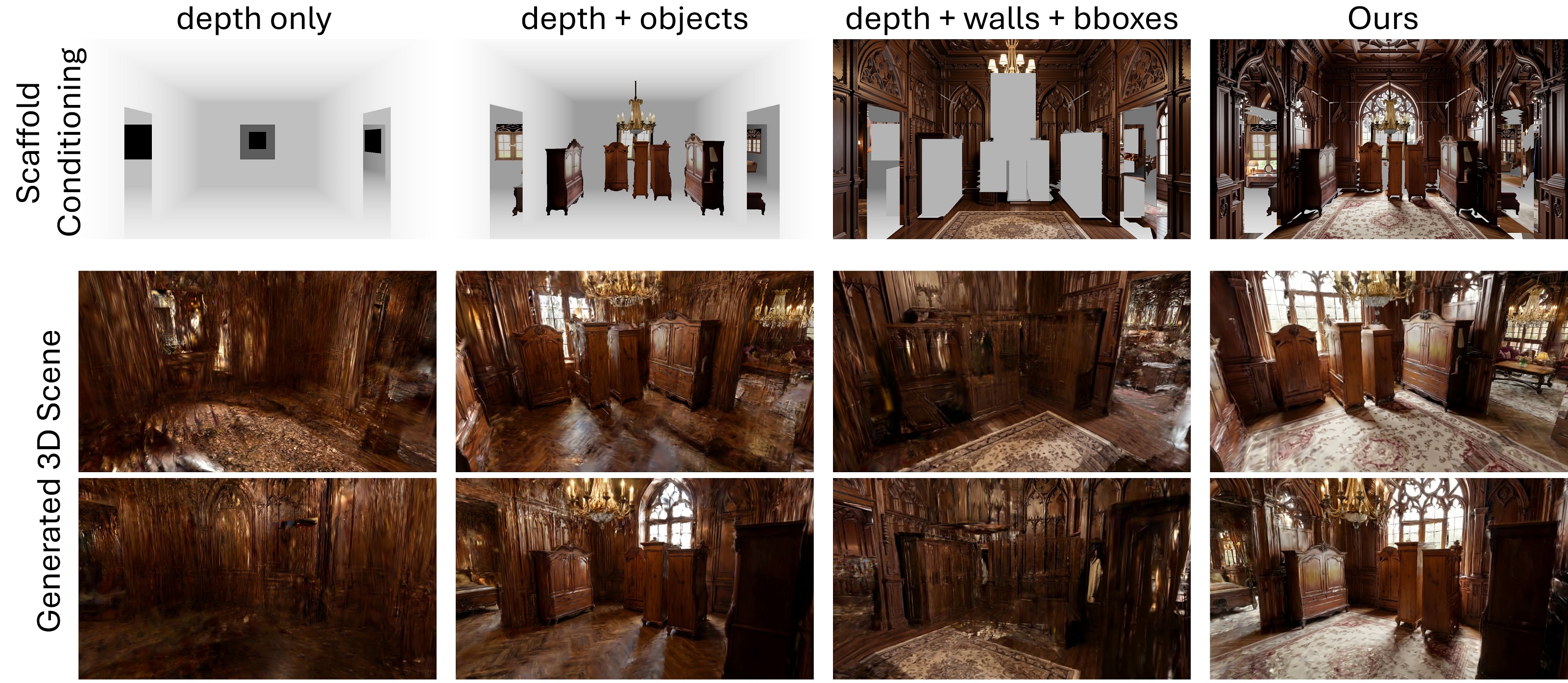}
\end{center}
\vspace{-0.5cm}
\caption{Ablation of mesh scaffold conditioning. Using only the structural mesh without  wall textures (\emph{depth only}) produces large inconsistencies at the local and object level. Using our scaffold mesh with objects but without wall textures (\emph{depth+objects}) preserves object coherency, but still exhibits inconsistencies on structural elements. Structural conditioning with wall texture, but with objects represented as bounding boxes (\emph{depth+walls+bboxes}), does not sufficiently constrain 3D consistency. Our full mesh scaffold with wall texture achieves strong local and global 3D consistency.
}
\label{fig:ablation}
\end{figure}

\subsection{Quantitative Results}
\label{sec:quantitative}

We report automatic image quality metrics and a perceptual study in \cref{tab:comparison}.
CLIP-IQA+~\cite{wang2022exploring} measures technical image quality (sharpness, noise, artifacts), while CLIP Aesthetic~\cite{schuhmann2022clip} evaluates subjective visual appeal (composition, color harmony).
Our method leads on both metrics; however, these only assess 2D quality and do not capture 3D consistency. 

The perceptual study highlights 3D consistency, with \OURS{} outperforming all baselines across all three dimensions: object consistency across viewpoints (\emph{3D Objects}), spatial coherence of room geometry (\emph{3D Structure}), and overall visual fidelity and 3D coherence (\emph{Quality}). Pairwise comparisons further confirm this, with users preferring our results in 96.2\% of cases.
This demonstrates that our mesh scaffold provides a superior level of 3D consistency.

\begin{table}[t]
  \caption{Quantitative comparison with state of the art. Perceptual study ratings (1--5 scale, higher is better) across 31 participants.  Pairwise preferences (right) from perceptual study show \% of participants preferring \OURS{} over each baseline. }
  \vspace{-0.2cm}
  \label{tab:comparison}
\begin{minipage}{0.65\textwidth}
  \centering
  \setlength{\tabcolsep}{5pt}
  \resizebox{\linewidth}{!}{
  \begin{tabular}{l cc ccc}
    \toprule
    & \multicolumn{2}{c}{Automatic Metrics $\uparrow$} & \multicolumn{3}{c}{User Study $\uparrow$} \\
    \cmidrule(lr){2-3} \cmidrule(lr){4-6}
    Method & IQA+ & Aesth. & Quality & 3D Objects & 3D Structure \\
    \midrule
    WonderWorld & 0.4872 & 4.7795 & 1.94 & 2.19 & 1.84 \\
    FlexWorld & 0.4097 & 4.9453 & 2.05 & 2.00 & 2.34 \\
    LayerPano3D & 0.3949 & 4.7312 & 2.27 & 2.21 & 2.44 \\
    WorldExplorer & 0.4053 & 5.5476 & 2.58 & 2.47 & 2.76 \\
    SpatialGen & 0.4648 & 5.0633 & 2.61 & 3.00 & 3.00 \\
    DreamScene360 & 0.3565 & 4.7270 & 3.19 & 3.10 & 3.37 \\
    \textbf{Ours} & \textbf{0.5114} & \textbf{5.5799} & \textbf{4.48} & \textbf{4.40} & \textbf{4.35} \\
    \bottomrule
  \end{tabular}
  }
  \end{minipage}
  \begin{minipage}{0.3\textwidth}
  \centering
  \resizebox{\linewidth}{!}{
  \begin{tabular}{l c}
    \toprule
    Ours vs. & Preference (\%) \\
    \midrule
    SpatialGen & 100.0 \\
    WonderWorld & 100.0 \\
    WorldExplorer & 96.8 \\
    LayerPano3D & 93.5 \\
    DreamScene360 & 93.5 \\
    FlexWorld & 93.5 \\
    \midrule
    Average & 96.2 \\
    \bottomrule
  \end{tabular}
  }
  \end{minipage}
\end{table}

\begin{table}[tb]
  \centering
  \caption{\CR{Ablation study. \textbf{(a)}~Automatic metrics over five multi-room
  scenes: cross-view depth reprojection (MAE-norm, AbsRel), mesh-depth/normal alignment of the
  final 3DGS to our scaffold (\texttt{M-}/\texttt{N-}, \emph{Ours} only), and CLIP-IQA+/Aesthetic
  (separate eval set, so IQA+/Aesth.\ differ slightly from \cref{tab:comparison}).
  \textbf{(b)}~Perceptual study (1--5, 31 participants) on the gothic-revival scene, for the
  modality variants (depth+objects\,=\,w/o wall texture, depth+walls+bboxes\,=\,w/o object
  recon., depth only\,=\,w/o objects or wall texture). \textbf{Bold}\,=\,best.}}
  \label{tab:ablation}
  \vspace{-0.3cm}
  \scriptsize\setlength{\tabcolsep}{3pt}
  \begin{minipage}{\textwidth}\centering
  \textbf{(a) 3D consistency and image quality (automatic, 5 scenes)}\par\vspace{2pt}
  \resizebox{\textwidth}{!}{%
  \begin{tabular}{l cc cccc cc}
    \toprule
    & \multicolumn{2}{c}{3D cons.\ (Reproj.)} & \multicolumn{4}{c}{Mesh alignment (\emph{Ours} only)} & \multicolumn{2}{c}{Quality} \\
    \cmidrule(lr){2-3}\cmidrule(lr){4-7}\cmidrule(lr){8-9}
    Variant & MAE-norm\,$\downarrow$ & AbsRel\,$\downarrow$ & M-MAE\,$\downarrow$ & M-RMSE\,$\downarrow$ & N-Cos\,$\uparrow$ & N-Ang\,$\downarrow$ & IQA+\,$\uparrow$ & Aesth.\,$\uparrow$ \\
    \midrule
    \textbf{Full (Ours, NB Pro)} & \textbf{0.1615} & \textbf{0.0761} & \textbf{0.0383} & \textbf{0.1285} & \textbf{0.6893} & \textbf{39.26} & \textbf{0.5529} & \textbf{5.5925} \\
    \midrule
    \quad depth+objects      & 0.2229 & 0.0997 & 0.1167 & 0.2485 & 0.4033 & 62.01 & 0.3870 & 5.1057 \\
    \quad depth+walls+bboxes & 0.2722 & 0.1988 & 0.3335 & 0.6387 & 0.3452 & 66.17 & 0.3735 & 5.2041 \\
    \quad depth only         & 0.1970 & 0.0888 & 0.2777 & 0.5857 & 0.3780 & 63.86 & 0.3818 & 5.0515 \\
    \midrule
    \quad \CR{w/o edge recall validation} & \CR{0.1660} & \CR{0.0784} & \CR{0.0478} & \CR{0.1437} & \CR{0.6489} & \CR{42.65} & \CR{0.5237} & \CR{5.5482} \\
    \quad \CR{w/o mesh 3DGS pcd init}     & \CR{0.1706} & \CR{0.0803} & \CR{0.0479} & \CR{0.1406} & \CR{0.6433} & \CR{42.93} & \CR{0.5204} & \CR{5.5831} \\
    \quad \CR{w/o mesh 3DGS depth loss}   & \CR{0.3009} & \CR{0.1641} & \CR{0.1904} & \CR{0.3969} & \CR{0.4574} & \CR{57.72} & \CR{0.5312} & \CR{5.5573} \\
    \quad \CR{w/o mesh 3DGS init / depth} & \CR{1.8804} & \CR{0.6730} & \CR{2.2549} & \CR{4.6730} & \CR{0.2205} & \CR{74.65} & \CR{0.4393} & \CR{5.4270} \\
    \bottomrule
  \end{tabular}}
  \end{minipage}\par\vspace{3pt}
  \begin{minipage}{\textwidth}\centering
  \textbf{(b) Perceptual study (1--5 scale, 31 participants, single scene)}\par\vspace{2pt}
  \resizebox{0.62\textwidth}{!}{%
  \begin{tabular}{l ccc}
    \toprule
    Variant & Quality\,$\uparrow$ & 3D Objects\,$\uparrow$ & 3D Structure\,$\uparrow$ \\
    \midrule
    \textbf{Full (Ours)}     & \textbf{4.48} & \textbf{4.40} & \textbf{4.35} \\
    \midrule
    \quad depth+objects      & 3.35 & 3.29 & 3.52 \\
    \quad depth+walls+bboxes & 2.19 & 1.94 & 2.26 \\
    \quad depth only         & 2.10 & 1.97 & 2.26 \\
    \bottomrule
  \end{tabular}}
  \end{minipage}
\end{table}

\CR{We further report per-scene scaling and cost, a navigability analysis, and a 3D-consistency comparison across baselines and image backbones in the supplementary material.}

\subsection{Ablation Studies}
\label{sec:ablation}

We ablate both the image-conditioning modalities and the 3DGS optimization and validation
components in \Cref{tab:ablation}. \CR{Panel~(a) reports automatic 3D-consistency metrics over
five multi-room scenes, while panel~(b)} reports the perceptual study on the same six-room
``gothic revival mansion with ornate detail and dark wood'' scene, whose conditioning inputs
and output frames are shown in \Cref{fig:ablation}.

\paragraph{Conditioning modalities.}
Using only grayscale depth (\emph{depth only}) yields photorealistic images but inconsistent, ghosted furniture across views. Adding textured walls with objects as gray bounding boxes (\emph{depth+walls+bboxes}) improves surface-color consistency but leaves objects unconstrained, whereas reconstructed objects with untextured walls (\emph{depth+objects}) anchor furniture geometry yet force per-frame hallucination of wall appearance. Combining all signals (depth, textured walls, and reconstructed objects) gives comprehensive guidance for consistent generation, also reflected in the lowest reprojection and mesh-alignment errors in panel~(a).

\CR{\paragraph{3DGS optimization and validation.}
The mesh-conditioned 3DGS initialization and depth loss dominate 3D consistency: removing both collapses reprojection accuracy (MAE-norm $1.88$ vs.\ $0.16$) while image quality stays largely intact, isolating the consistency gain of mesh conditioning from raw appearance. The edge-recall validation and the point-cloud initialization each add smaller but consistent gains.}

\paragraph{Limitations.}
Our method is currently limited to single-story layouts and does not directly handle multi-level environments, though these could be generated floor by floor with more cohesive layout conditioning. It also relies on SAM-3D-Objects~\cite{meta2025sam3dobjects} for object reconstruction, which can leave incomplete back faces or missing detail in occluded regions.

\section{Conclusion}
\label{sec:conclusion}

We presented \OURS{}, an approach for generating navigable multi-room 3D
scenes from text by decoupling global spatial consistency from local
photorealistic appearance. We first construct an explicit 3D mesh scaffold that captures the scene’s geometric layout and maintains spatial coherence across rooms. Image diffusion is then conditioned on this scaffold (through its depth, object geometry, and progressively accumulated wall textures) to synthesize realistic appearance, with a depth-based validation loop ensuring structural fidelity.
Our mesh-anchored generation supports consistent appearance across diverse viewpoints, from close-up object views to long-range views spanning multiple rooms, opening a promising path towards truly environment-scale 3D world synthesis.

\section*{Acknowledgements} 
\begin{sloppypar}
This work has been supported by the ERC Starting Grant SpatialSem (101076253).
\end{sloppypar}

\clearpage

\bibliographystyle{splncs04}
\bibliography{main}

\clearpage
\appendix
\setcounter{figure}{0}
\renewcommand{\thefigure}{A\arabic{figure}}
\setcounter{table}{0}
\renewcommand{\thetable}{A\arabic{table}}

\section{Video}

Please watch our  video for additional results, as well as a video comparison of our method. We compare against baselines and ablations. We also provide a diverse set of scene layouts and themes rendered at different trajectories to showcase the quality of our outputs and the robustness of our method.

\section{Additional Results}
\begin{figure}[tp]
\begin{center}
\includegraphics[width=0.95\textwidth]{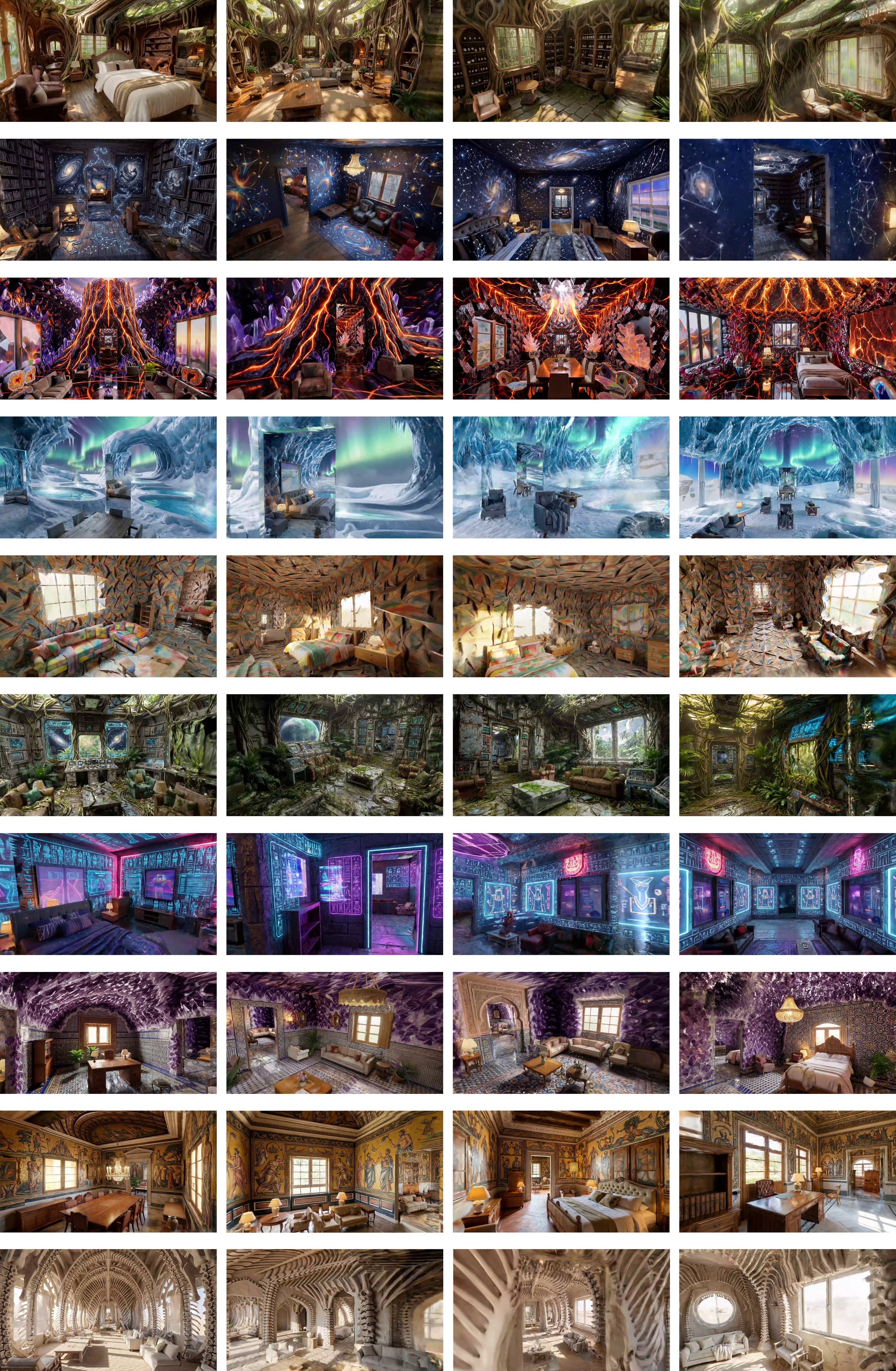}
\end{center}
\vspace{-0.5cm}
\caption{\OURS{} can synthesize various large, multi-room spaces with varying layouts and diverse visual atmospheres. Each row shows views from a generated 3D world.
}
\label{fig:resultsmore}
\end{figure}

We show additional scenes generated by \OURS{} in \cref{fig:resultsmore}, highlighting a diverse array of scene arrangement and visual themes. 

\begin{figure}[tp]
\begin{center}
\includegraphics[width=0.9\textwidth]{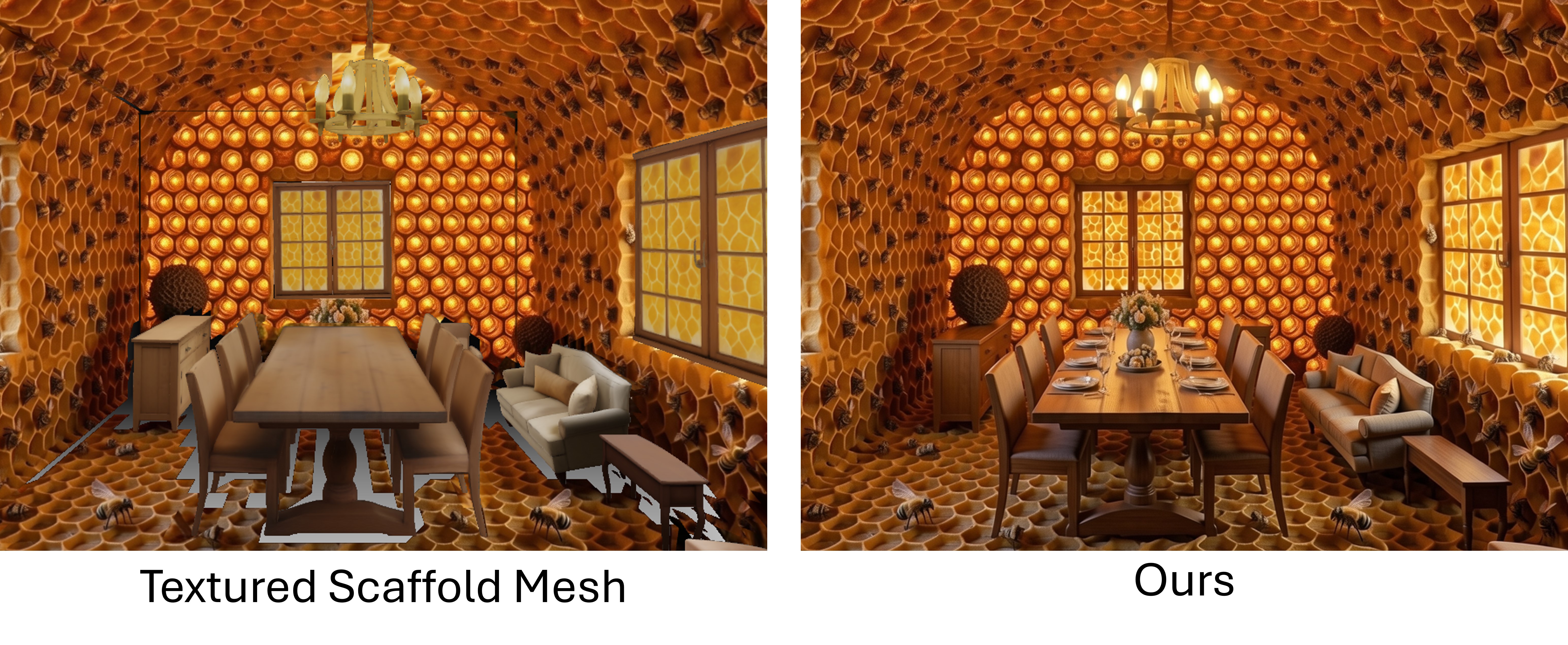}
\end{center}
\caption{Comparison of textured scaffold mesh with our final synthesized \CR{appearance}. The textured scaffold mesh provides a good initialization to encourage multi-view consistent synthesis, but lacks completeness, scene cohesiveness (e.g., seams along wall boundaries, less consistent object appearance between objects), and  high-fidelity detail which we provide through our final appearance synthesis. 
}
\label{fig:scaffoldtexture}
\end{figure}

Additionally, we show a visual comparison of our textured scaffold mesh to our final \CR{synthesized} scene appearance in \cref{fig:scaffoldtexture}, highlighting the impact of our mesh-conditioned appearance synthesis in producing complete, coherent large-scale 3D worlds with rich, high-fidelity detail.

\section{Additional Implementation Details}

\paragraph{Scaffold Mesh Texturing Details}
Multi-view consistency of 3D objects is ensured through their reconstruction and texturing during the mesh creation phase. To ensure that background information stays similarly consistent, we employ an iterative mesh texturing approach for the room walls based on previously generated images.

To ensure high-quality projections, we apply several geometric filters.
A camera-facing constraint excludes back-facing surfaces using the normal–view direction dot product. An occlusion filter compares projected vertex depth with the rendered scene depth (tolerance $\tau = 0.10$\,m) to prevent projecting foreground object pixels onto background walls. Room-based filtering restricts updates to surfaces belonging to the current room, preventing cross-room texture leakage.

Texture accumulation is progressive: each validated camera view contributes additional coverage to previously untextured surfaces. Over time, this produces an increasingly complete and globally consistent wall texture map directly embedded into the final output $\mathcal{M}$. Importantly, this consistency is enforced at the geometric level, independent of diffusion, allowing future views to inherit accumulated appearance information from the scaffold itself.

The textured mesh in \cref{fig:scaffoldtexture} is the final textured mesh after all images have been generated. Throughout the pipeline the textured mesh is therefore not only input but also output to the image generation model.

\paragraph{Object Segmentation.}

We use SAM3~\cite{meta2025sam3} in its interactive point-based mode to segment individual objects from the generated images. For each room view, the  user provides positive and negative point prompts on the image via a lightweight annotation interface; SAM3 returns a mask prediction after each click, enabling iterative refinement through successive points. Once satisfactory, the mask is confirmed and assigned an object label (e.g., ``sofa'', ``lamp''). The resulting per-object binary masks undergo post-processing to discard small detections, merge overlapping regions corresponding to the same object, and split disconnected components into separate instances.

This annotation step can alternatively be automated using a vision-language
model to generate text prompts that drive SAM3 in its text-prompted mode, removing the need for manual interaction.

\paragraph{Object Placement.}

Each object reconstructed by SAM-3D-Objects~\cite{meta2025sam3dobjects} is produced in camera space. 
The SAM-3D-Objects pose (rotation and translation) is applied to
bring the object into camera coordinates, which are then transformed to world coordinates
using the known camera-to-world extrinsics.
Once in world space, each object is classified as floor-standing, flat (e.g., rug),
wall-mounted, or ceiling-hung based on its label and aspect ratio.
Floor-standing objects are leveled by computing their oriented bounding box (OBB),
identifying the face most aligned with the ground plane, and rotating the mesh
so that face becomes horizontal. They are then dropped to the floor or stacked
on previously placed objects using axis-aligned overlap detection.
Wall-mounted and ceiling objects are positioned at the appropriate height.
Finally, XY conflicts between objects are resolved by shifting overlapping
items toward nearby walls while preserving their relative spatial arrangement.
All placed objects are merged with the structural room mesh into a single scene.

\paragraph{Iterative Generation Prompt}                                                                                                                                                                                                                                           
During the final image generation stage of our method, each camera view (except the first bootstrap camera) is generated by conditioning on two inputs: (1) a 3D rendering from the target camera pose, which includes projected wall textures and reconstructed object meshes, and (2) the photorealistic output from a previous generation in the
sequence, which serves as a style reference. Both images are passed to Nano Banana Pro \cite{google2025nanobanana} alongside the text prompt shown in Figure~\ref{fig:iterative_prompt}, where \texttt{\{theme\}} is replaced with the scene description (e.g., ``a modern minimalist living room'').                             
                                                              
\begin{figure*}[p]
\centering
\noindent\fbox{\parbox{\dimexpr\textwidth-2\fboxsep-2\fboxrule}{%
\small\raggedright\ttfamily
The scene depicts a ``\{theme\}''

\medskip
The first image is a 3D rendering showing the correct camera angle, composition, and object layout for the desired output. However, it is not photorealistic---the objects are placeholders that show correct position, scale, and shape but not correct appearance. The rendering contains texture seams and stretching
artifacts from steep-angle projection that must be removed. Additionally, wall and surface textures in the rendering are flat 2D projections---the output should depict these surfaces with proper 3D depth, volume, and parallax-correct detail.

\medskip
The second image is a photorealistic photograph of the same scene from a very different camera angle. It defines how the scene should actually look.

\medskip
The camera positions differ significantly, so not all objects are visible in both images. Before generating, establish correspondences between objects in the rendering and the photograph based on shape, position, and spatial context.

\medskip
This image is part of an iterative generation pipeline, so the input rendering may contain accumulated visual artifacts such as blur, color degradation, loss of detail, or repeated compression artifacts from previous iterations. Actively counteract these by producing a clean, sharp, and visually pleasing
output---treat the generation as an opportunity to restore and enhance visual quality rather than propagate existing degradation.

\medskip
Generate a photorealistic image that:

1.~Preserves the exact camera angle, composition, and object layout from the rendering---do not add, remove, or rearrange any objects

2.~Transfers photorealistic appearance, materials, lighting, and textures from corresponding objects in the photograph---not from the rendering

3.~Does not introduce objects from the photograph that are absent in the rendering

4.~Removes all texture seams, stretching artifacts, flat 2D texturing, placeholder appearances, and any accumulated pipeline artifacts---surfaces should have realistic 3D depth and detail

5.~Is multi-view consistent with the photograph, as if both were real photos of the same physical scene

6.~For objects in the rendering with no match in the photograph, infers plausible photorealistic appearance consistent with the scene

\medskip
Use the rendering for geometry and camera only. Use the photograph for appearance only.
}}
\caption{Text prompt used for iterative multi-view generation in Stage~6. The model receives this prompt together with two images: a 3D rendering from the target camera pose and a photorealistic reference from a previous generation.}
\label{fig:iterative_prompt}
\end{figure*}

\section{Perceptual Study Details}

We conduct a perceptual study to evaluate the performance of \OURS{} and baselines. We ask 31 participants to rate the object consistency across view-points (3D Objects), spatial coherence of room geometry (3D Structure), and overall visual fidelity and 3D coherence (Quality) on short trajectory renderings of the final radiance-based outputs, with a particular focus on close-view trajectories and rotations around objects. Please refer to \CR{\cref{fig:userstudy}} for an example prompt of our perceptual study and exact question formulation.

\begin{center}
\includegraphics[width=0.62\textwidth]{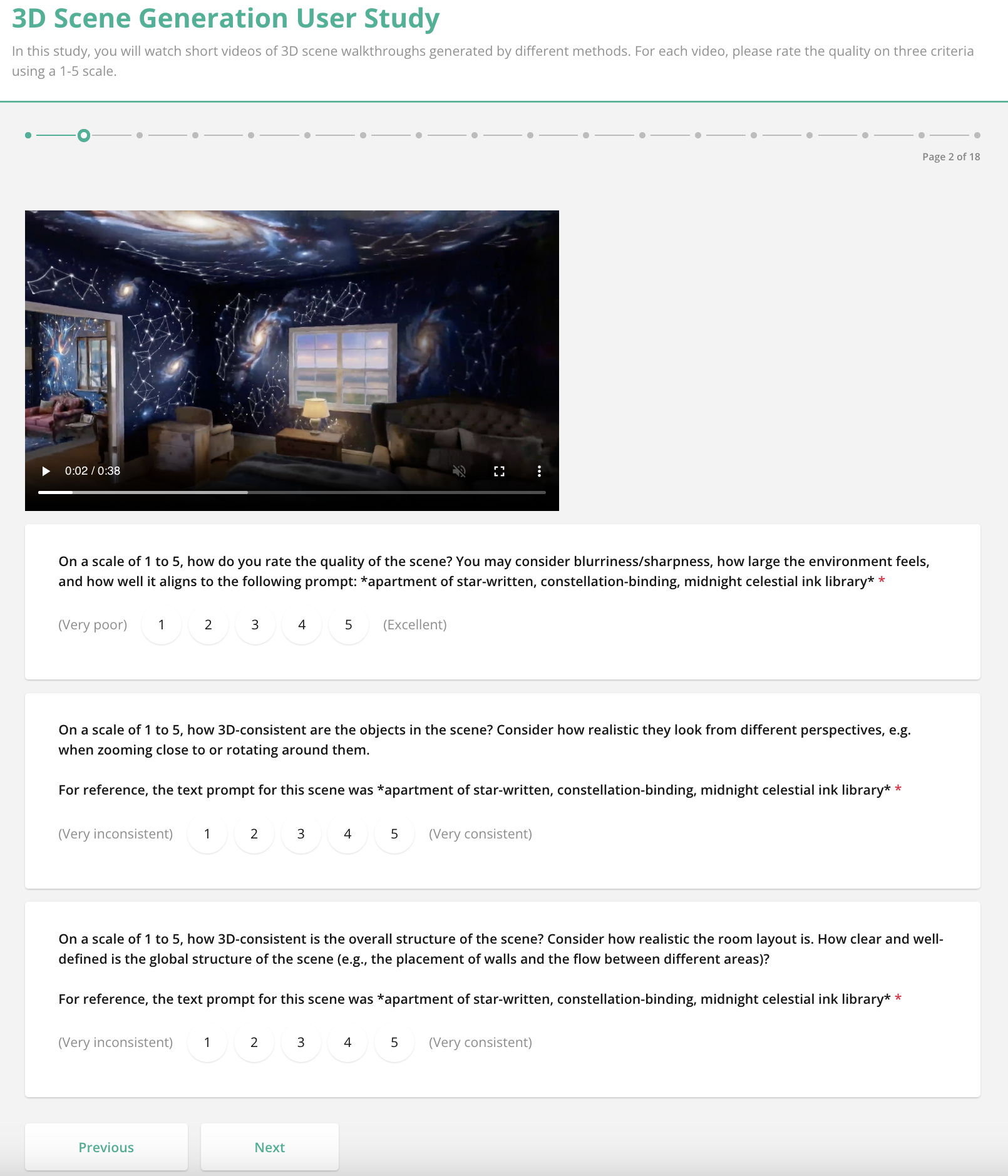}
\captionof{figure}{Example screenshot from our perceptual study. Participants are asked to rank consistency of 3D Objects and 3D Structure, as well as overall Scene Quality on a scale from 1 to 5.}
\label{fig:userstudy}
\end{center}
  
\section{Additional Baseline Details}

Here we provide some additional information on our baseline evaluation setup.

\emph{Layout-guided 3D Indoor Scene Generation.} We evaluate SpatialGen  \cite{SpatialGen} on our scenes by converting our textured mesh scaffolds into the required multi-view input format. For each room, we generate 16 views along a circular camera trajectory with a 90° field of view. Per view, we render: (1) an RGB image from the textured mesh, (2) a semantic layout image with per-category color coding (walls, floor, ceiling, and furniture classes), and (3) a layout depth map from the structural mesh. The first RGB view serves as the conditioning image, while the semantic layout and depth maps are used to construct the scene coordinate map that encodes the 3D layout prior. SpatialGen then generates 15 novel views conditioned on these inputs, which are reconstructed into a 3D scene via SpatialGen's custom Sparse-RaDeGS representation.  

\emph{Autoregressive Video Diffusion for Navigable 3D Scenes.} We evaluate FlexWorld \cite{chen2024flexworld} by providing one input image generated with the same theme text-prompt as used for our method as input, not changing any of the default hyperparameters of FlexWorld. We evaluate \CR{WorldExplorer} \cite{schneider2025worldexplorer} by providing the same theme text-prompt and keeping WorldExplorer's predefined trajectories for the scene generation. We then render out the resulting 3DGS scene on trajectories that focus on close-object rotations beyond the initial predefined trajectories used for scene generation, especially in regions not close to the scene center.

\emph{Iterative Text-to-Image Lifting.} To compare against WonderWorld \cite{yu2024wonderworld} we provide one image generated with the same theme text-prompt as is used in our method to WonderWorld. Subsequently we choose novel viewpoints following the manner outlined in the original WonderWorld work.

\emph{Panorama-To-3D.} Initialization of DreamScene360 \cite{zhou2024dreamscene360} and LayerPano3D \cite{yang2025layerpano3d} is only possible by providing panorama or text inputs. We compare to these baselines by providing the identical theme text-prompt that is used for our method.

\clearpage
\section{\texorpdfstring{\CR{Extended Quantitative Evaluation and Analysis}}{Extended Quantitative Evaluation and Analysis}}

\CR{This section collects the additional quantitative evaluation referenced in the main paper: a 3D-consistency comparison across methods and image backbones (\cref{tab:main}), per-scene scaling and cost (\cref{tab:scaling}), layout-LLM statistics (\cref{tab:llm-perf}), and a navigability visualization (\cref{fig:reachability}). Our method does not depend on any single foundation model. The corresponding component ablations, using the same 3D-consistency metrics, are reported in the main paper (\cref{tab:ablation}).}

\CR{\textbf{3D-consistency evaluation.} To quantify 3D consistency beyond the perceptual study, \cref{tab:main} reports, over five multi-room scenes, (i)~a cross-view depth-reprojection error that is comparable across methods (a per-method scale-normalized MAE, and the absolute relative error, AbsRel), and (ii)~depth and normal alignment of the final 3DGS to the mesh scaffold (\texttt{M-}/\texttt{N-}). Our method attains the lowest reprojection error of all methods, with both the local Flux2-klein and the NB~Pro backbone clearly outperforming the strongest baselines, confirming that our mesh scaffold, rather than any single image model, drives the gain in 3D consistency.}

\CR{\textbf{Scaling and cost.} The mesh scaffold also scales gracefully: per-scene cost from 4 to 12 rooms (\cref{tab:scaling}) grows roughly linearly while 3D-consistency and quality remain stable, with high acceptance (\emph{Yield}) at low average reseed counts (\emph{Atm}). With NB~Pro, rooms are generated in parallel; the fully-local Flux2-klein path runs sequentially on a single RTX~A5000.}

\CR{\textbf{Navigability.} Rasterizing each scene into a 2D occupancy grid (0.15\,m; walls and floor-resting objects as obstacles), the largest 4-connected free-space component covers \textbf{96.7\%} of free cells on average across the five scenes, confirming near-full traversability (\cref{fig:reachability}).}

\begin{center}
\begin{minipage}{\linewidth}\centering
  \scriptsize\setlength{\tabcolsep}{3pt}
  \captionof{table}{3D-consistency comparison across methods and image backbones over 5 scenes (4--8 rooms, different prompts). Reproj.\ metrics (cross-view depth reprojection: a per-method scale-normalized MAE, and AbsRel) are comparable across methods; \texttt{M-}/\texttt{N-} are mesh-depth/normal alignment of the final 3DGS to our scaffold (\emph{Ours} only). Computed on a separate evaluation set, so absolute IQA+/Aesth.\ differ slightly from \cref{tab:comparison}. The corresponding component ablations are reported in the main paper (\cref{tab:ablation}). \textbf{Bold}\,=\,best.}
  \label{tab:main}
  \resizebox{\textwidth}{!}{%
  \begin{tabular}{l cc cccc cc}
    \toprule
    & \multicolumn{2}{c}{3D cons.\ (Reproj.)} & \multicolumn{4}{c}{Mesh alignment (\emph{Ours} only)} & \multicolumn{2}{c}{Quality} \\
    \cmidrule(lr){2-3}\cmidrule(lr){4-7}\cmidrule(lr){8-9}
    Method & MAE-norm\,$\downarrow$ & AbsRel\,$\downarrow$ & M-MAE\,$\downarrow$ & M-RMSE\,$\downarrow$ & N-Cos\,$\uparrow$ & N-Ang\,$\downarrow$ & IQA+\,$\uparrow$ & Aesth.\,$\uparrow$ \\
    \midrule
    DreamScene360 & 0.4336 & 0.1282 & -- & -- & -- & -- & 0.3153 & 4.2699 \\
    WorldExplorer & 0.3221 & 0.1183 & -- & -- & -- & -- & 0.3633 & 5.5906 \\
    SpatialGen    & 0.4055 & 0.1825 & -- & -- & -- & -- & 0.4294 & 4.4503 \\
    \midrule
    Ours (full, Flux2-klein-9B) & 0.2173 & 0.0964 & 0.1002 & 0.2311 & 0.4716 & 56.74 & 0.4450 & 5.5384 \\
    \midrule
    Ours (full, NB Pro) & \textbf{0.1615} & \textbf{0.0761} & \textbf{0.0383} & \textbf{0.1285} & \textbf{0.6893} & \textbf{39.26} & \textbf{0.5529} & \textbf{5.5925} \\
    \bottomrule
  \end{tabular}}
\end{minipage}
\end{center}

\begin{center}
\includegraphics[width=0.47\textwidth]{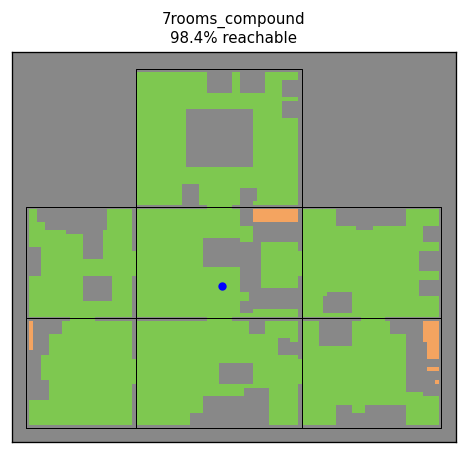}
\captionof{figure}{Navigability. A generated layout rasterized into a 2D occupancy grid (0.15\,m): green\,=\,reachable free space, gray\,=\,obstacles (walls/objects), orange\,=\,unreachable free space, dot\,=\,seed. Free cells in the largest 4-connected component average 96.7\% over our five scenes.}
\label{fig:reachability}
\end{center}

\begin{center}
\begin{minipage}{\linewidth}\centering
  \small\setlength{\tabcolsep}{6pt}
  \captionof{table}{Floor-plan layout-LLM backends: mean correction iterations and wall-clock until a spatially-valid layout, over 10 runs of varying scene size. All backends reach valid layouts within a few iterations.}
  \label{tab:llm-perf}
  \begin{tabular}{@{}lcc@{}}
    \toprule
    Model & Mean iters & Time \\
    \midrule
    Opus 4.6~\cite{anthropic2026opus46}                          & 1.30 & 79\,s \\
    Qwen2.5-Coder-32B~\cite{hui2024qwencoder} (local, RTX A5000) & 1.40 & 252\,s \\
    gpt-oss-20b~\cite{openai2025gptoss} (local, RTX A5000)       & 2.60 & 229\,s \\
    Haiku 4.5~\cite{anthropic2025haiku45}                        & 3.20 & 101\,s \\
    \bottomrule
  \end{tabular}
\end{minipage}
\end{center}

\begin{center}
\begin{minipage}{\linewidth}\centering
  \scriptsize\setlength{\tabcolsep}{3pt}
  \captionof{table}{Per-scene scaling and cost. Each cell: \textit{Ours--NB Pro} / \textit{Ours--Flux2-klein-9B} (distilled, 4 sampling steps). \textbf{Atm}: mean reseed attempts per accepted view; \textbf{Yield}: \% of intended views retained; \textbf{Time}: end-to-end wall-clock per scene. NB~Pro is accessed through its API, which lets us generate rooms in parallel, whereas the fully-local Flux2-klein-9B path runs sequentially on a single RTX~A5000. 3D-consistency and quality remain stable across room counts and backbones.}
  \label{tab:scaling}
  \resizebox{\textwidth}{!}{%
  \begin{tabular}{c cc cccc cc ccc}
    \toprule
    \#Rooms & Reproj-MAE-norm\,$\downarrow$ & Reproj-AbsRel\,$\downarrow$ & M-MAE\,$\downarrow$ & M-RMSE\,$\downarrow$ & N-Cos\,$\uparrow$ & N-Ang\,$\downarrow$ & IQA+\,$\uparrow$ & Aesth.\,$\uparrow$ & Atm\,$\downarrow$ & Yield\,$\uparrow$ & Time\,$\downarrow$ \\
    \midrule
    4 (80\,m\textsuperscript{2})   & 0.11 / 0.17 & 0.07 / 0.10 & 0.02 / 0.07 & 0.07 / 0.15 & 0.75 / 0.50 & 34.72 / 54.50 & 0.49 / 0.44 & 5.56 / 5.74 & 1.20 / 1.68 & 74\% / 74\% & 0.8\,h / 2.2\,h \\
    6 (144\,m\textsuperscript{2})  & 0.19 / 0.23 & 0.10 / 0.11 & 0.05 / 0.11 & 0.18 / 0.29 & 0.61 / 0.39 & 45.93 / 62.59 & 0.54 / 0.39 & 5.54 / 5.52 & 1.22 / 1.35 & 86\% / 72\% & 1.0\,h / 3.3\,h \\
    8 (192\,m\textsuperscript{2})  & 0.15 / 0.25 & 0.05 / 0.08 & 0.05 / 0.14 & 0.15 / 0.29 & 0.63 / 0.44 & 44.59 / 59.21 & 0.62 / 0.49 & 5.56 / 5.27 & 1.24 / 1.56 & 71\% / 75\% & 1.2\,h / 4.4\,h \\
    12 (256\,m\textsuperscript{2}) & 0.23 / 0.33 & 0.08 / 0.11 & 0.05 / 0.13 & 0.16 / 0.31 & 0.61 / 0.43 & 45.88 / 59.76 & 0.54 / 0.36 & 5.55 / 5.54 & 1.24 / 1.51 & 82\% / 79\% & 1.5\,h / 6.6\,h \\
    \bottomrule
  \end{tabular}}
\end{minipage}
\end{center}

\end{document}